\documentclass[10pt,twocolumn,letterpaper]{article}

\usepackage{iccv}
\usepackage{times}
\usepackage{epsfig}
\usepackage{graphicx}
\usepackage{amsmath}
\usepackage{amssymb}
\usepackage{subcaption}
\usepackage{flushend}
\usepackage{bm}

\usepackage[pagebackref=true,breaklinks=true,letterpaper=true,colorlinks,bookmarks=false]{hyperref}

\iccvfinalcopy 

\def\iccvPaperID{5184} 

\ificcvfinal\fi

\begin{document}

\title{Watch, Listen and Tell: Multi-modal Weakly Supervised Dense Event Captioning}

\author{Tanzila Rahman$^{1,2}$ \qquad Bicheng Xu$^{1,2}$ \qquad Leonid Sigal$^{1,2,3}$\\
$^1$University of British Columbia \qquad
$^2$Vector Institute for AI \qquad
$^3$Canada CIFAR AI Chair \\
{\tt\small \{trahman8, bichengx, lsigal\}@cs.ubc.ca}
}


\maketitle
\ificcvfinal\thispagestyle{empty}\fi

\begin{abstract}
Multi-modal learning, particularly among imaging and linguistic modalities, has made amazing strides in many high-level fundamental visual understanding problems, ranging from language grounding to dense event captioning. However, much of the research has been limited to approaches that either do not take audio corresponding to video into account at all, or those that model the audio-visual correlations in service of sound or sound source localization. In this paper, we present the evidence, 
that audio signals can carry surprising amount of information when it comes to high-level visual-lingual tasks. Specifically, we focus on the problem of weakly-supervised dense event captioning in videos and show that audio on its own can nearly rival performance of a state-of-the-art visual model and, combined with video, can improve on the state-of-the-art performance. Extensive experiments on the ActivityNet Captions dataset show that our proposed multi-modal approach outperforms state-of-the-art unimodal methods, as well as validate specific feature representation and architecture design choices. 
\end{abstract}

\vspace{-0.20in}
\section{Introduction}
Humans often perceive the world through multiple sensory modalities, such as watching, listening, smelling, touching, and tasting. 
Consider two people sitting in a restaurant; seeing them across the table suggests that they maybe friends or coincidental companions; hearing, even the coarse demeanor of their conversation, makes the nature of their relationship much clearer.
In our daily life, there are many other examples that produce strong evidence that multi-modal co-occurrences give us fuller perception of events. 
Recall how difficult it is to perceive the intricacies of the story from a {\em silent} film. 
Multi-modal perception has been widely studied in areas like psychology~\cite{davenport1973cross,vroomen2000sound}, neurology~\cite{stein1993merging}, and human computer interaction~\cite{tanveer2015unsupervised}. 

In the computer vision community, however, the progress in learning representations from multiple modalities has been limited, especially for high-level perceptual tasks where such modalities (\eg, audio or sound) can play an integral role.  
Recent works~\cite{owens2018audio,senocak2018learning} propose approaches for localizing audio in unconstrained videos (sound source localization) or utilize sound in video captioning~\cite{hao2018integrating, hori2017attention, wang2018watch, tian2018attempt}. 
However, these approaches consider relatively short videos, \ie, usually about 20 seconds, and focus on description of a single salient event~\cite{xu2016msr}. 
More importantly, while they show that audio can boost the performance of visual models to an extent, such improvements are typically considered marginal and the role of audio is delegated to being secondary (or not nearly as important) as visual signal~\cite{hori2017attention,wang2018watch}. 

\begin{figure}
  \centering
  \includegraphics[scale=0.50]{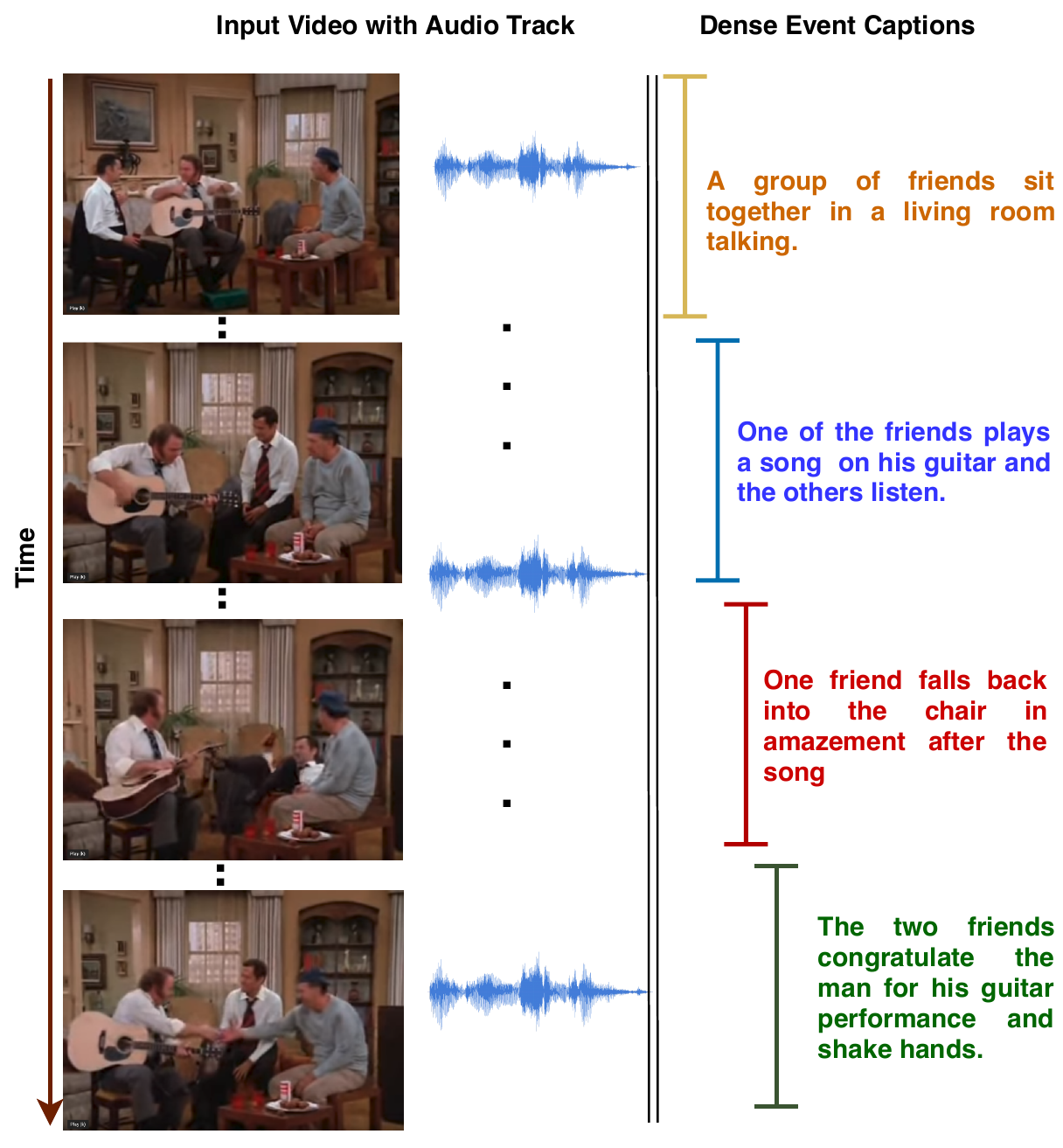}
\caption{
{\bf Multi-modal Dense Event Captioning.}  Illustration of our problem definition, where we use both audio features and visual information to generate the dense captions for a video in a weakly supervised manner.}
\label{fig1:intro}
\vspace{-0.25in}
\end{figure}


We posit that sound (or audio) may in fact be much more important than the community may realize. Consider the previously mentioned example of a silent film. The lack of sound makes it significantly more difficult, if not impossible in many cases, to describe the rich flow of the story and constituent events. Armed with this intuition, we focus on dense event captioning \cite{li2018jointly,wang2018bidirectional,zhou2018end} ({\em a.k.a.} dense-captioning of events in videos~\cite{krishna2017dense}) and endow our models with ability to utilize rich auditory signals for both event localization and captioning. Figure~\ref{fig1:intro} illustrates one example of our multi-modal dense event captioning task.
Compared with conventional video captioning, dense event captioning deals with longer and more complex video sequences, usually 2 minutes or more.
To the best of our knowledge, our work is the first to tackle the dense event captioning with sound, treating sound as a first class perceptual modality.

Audio features can be represented in many different ways. Choosing the most appropriate representation for our task is challenging. To this end, we compare different audio feature representations in this work. 
Importantly, we show that audio signal alone can achieve impressive performance on the dense event captioning task (rivalling visual counterpart). 
The form of fusion needed to incorporate the audio with the video signal is another challenge. 
We consider and compare a variety of fusion strategies. 

Dense event captioning provides detailed descriptions for videos, which is beneficial for in-depth video analysis. However, training a fully supervised model requires both caption annotations and corresponding temporal segment coordinates (\ie, the start and end time of each event), which is extremely difficult and time consuming to collect. Recently, \cite{duan2018weakly} proposes a method for dense event captioning in a weakly supervised setting. The approach does not require temporal segment annotation during training. During evaluation, the model is able to detect all events of interest and generate their corresponding captions. Inspired by and building on~\cite{duan2018weakly}, we tackle our multi-modal dense event captioning in a weakly supervised manner.

\vspace{0.05in}
\noindent
{\bf Contributions.} Our contributions are multiple fold. First, to the best of our knowledge, this is the first work that addresses dense event captioning task in a multi-modal setting. In doing so, we propose an attention-based multi-modal fusion model to integrate both audio and video information. Second, we compare different audio feature extraction techniques~\cite{aytar2016soundnet, davis1980comparison, lidy2016cqt}, and analyze their suitability for the task.
Third, we discuss and test different fusion strategies for incorporating audio cues with visual features. Finally, extensive experiments on the ActivityNet Captions dataset~\cite{krishna2017dense} show that audio model, on its own, can nearly rival performance of a visual model and, combined with video, using our multi-modal weakly-supervised approach, can improve on the state-of-the-art performance.

\section{Related Work}

\noindent
{\bf Audio Feature Representations.}
Recently computer vision community has begun to explore audio features for learning good representations in unconstrained videos. Aytar \etal ~\cite{aytar2016soundnet} propose a sound network guided by a visual teacher to learn the representations for sound. 
Earlier works, \cite{owens2018audio, senocak2018learning, sun2018indoor}, address sound source localization problem to identify which pixels or regions are responsible for generating a specified sound in videos (sound grounding). For example, \cite{senocak2018learning}~introduces an attention based localization network guided by sound information. A joint representation between audio and visual networks is presented in~\cite{owens2018audio, sun2018indoor} to localize sound source. Gao \etal~\cite{gao2018learning} formulate a new problem of audio source separation using a multi-instance multi-label learning framework. This framework maps audio bases, extracted by non-negative matrix factorization (NMF), to the detected visual objects. In recent year, audio event detection (AED)~\cite{cakir2015polyphonic, parascandolo2016recurrent, takahashi2016deep} has received attention in the research community. Most of the AED methods locate audio events and then classify each event.

\vspace{0.05in}
\noindent
{\bf Multi-modal Features in Video Analysis.}
Combining audio with visual features (\ie, multi-modal representation) often boosts performance of networks in vision, especially in video analysis ~\cite{arabaci2018multi, ariav2019end,hori2017attention,tian2018attempt,wang2018watch}. Ariav \etal~\cite{ariav2019end} propose an end-to-end deep neural network to detect voice activity by incorporating audio and visual modalities. Features from both modalities are fused using multi-modal compact bilinear pooling (MCB) to generate a joint representation for speech signal. 
Authors in~\cite{arabaci2018multi} propose a multi-modal method for egocentric activity recognition where audio-visual features are combined with multi-kernel learning and boosting. 

Recently, multi-modal approaches are also gaining popularity for video captioning~\cite{tian2018attempt,wang2018watch}. In~\cite{hori2017attention} a multi-modal attention mechanism to fuse information across different modalities is proposed. 
Hori~\etal~\cite{hori2017early} extend the work in~\cite{hori2017attention} by applying hypothesis-level integration based on minimum Bayes-risk decoding~\cite{kumar2004minimum,stolcke1997explicit} to improve the caption quality. 
Hao \etal~\cite{hao2018integrating} present multi-modal feature fusion strategies to maximize the benefits of visual-audio resonance information. 
Wang \etal~\cite{wang2018watch} introduce a hierarchical encoder-decoder network to adaptively learn the attentive representations of multiple modalities, and fuse both global and local contexts of each modality for video understanding and sentence generation. A 
module for exploring modality selection during sentence generation is proposed in~\cite{tian2018attempt} with the aim to interpret how words in the generated sentences are associated with audio and visual modalities. 

\vspace{0.05in}
\noindent
{\bf Dense Event Captioning in Videos.}
The task of dense event captioning in videos was first introduced in~\cite{krishna2017dense}. The task involves detecting multiple events that occur in a video and describing each event using natural language. 
Most of the works~\cite{liu2018best,yao2017msr} solve this problem in a two-stage manner, \ie, first temporal event proposal generation and then sentence captioning for each of the proposed event segments. In~\cite{yao2017msr}, authors adopt a temporal action proposal network to localize proposals of interest in videos, and then generate descriptions for each proposal. 
Wang \etal~\cite{wang2018bidirectional} present a bidirectional proposal method that effectively exploits both past and future contexts to make proposal predictions. 
In~\cite{zhou2018end}, a differentiable masking scheme is used to ensure the consistency between proposal and captioning modules. Li \etal~\cite{li2018jointly} propose a descriptiveness regression component to unify the event localization and sentence generation. 
Xu \etal~\cite{xu2019joint} present an end-to-end joint event detection and description network (JEDDi-Net) which adopts region convolutional 3D network~\cite{xu2017r} for proposal generation and refinement, and proposes hierarchical captioning.

Duan \etal~\cite{duan2018weakly} formulate the dense event captioning task in a weakly supervised setting, where there is no ground-truth temporal segment annotations during training and evaluation. They decompose the task into a pair of dual problems, event captioning and sentence localization, and present an iterative approach for training. Our work is motivated by~\cite{duan2018weakly} and builds on their framework. However, importantly, we fuse audio and visual features, and explore a variety of fusion mechanisms to address the multi-modal weakly supervised dense event captioning task. We note that~\cite{duan2018weakly} is thus far the only method for dense event captioning in the weakly supervised setting.    
\vspace{-0.10in}
\section{Multi-modal Dense Event Captioning}
\begin{figure*}[ht]
  \centering
  \includegraphics[scale=0.70]{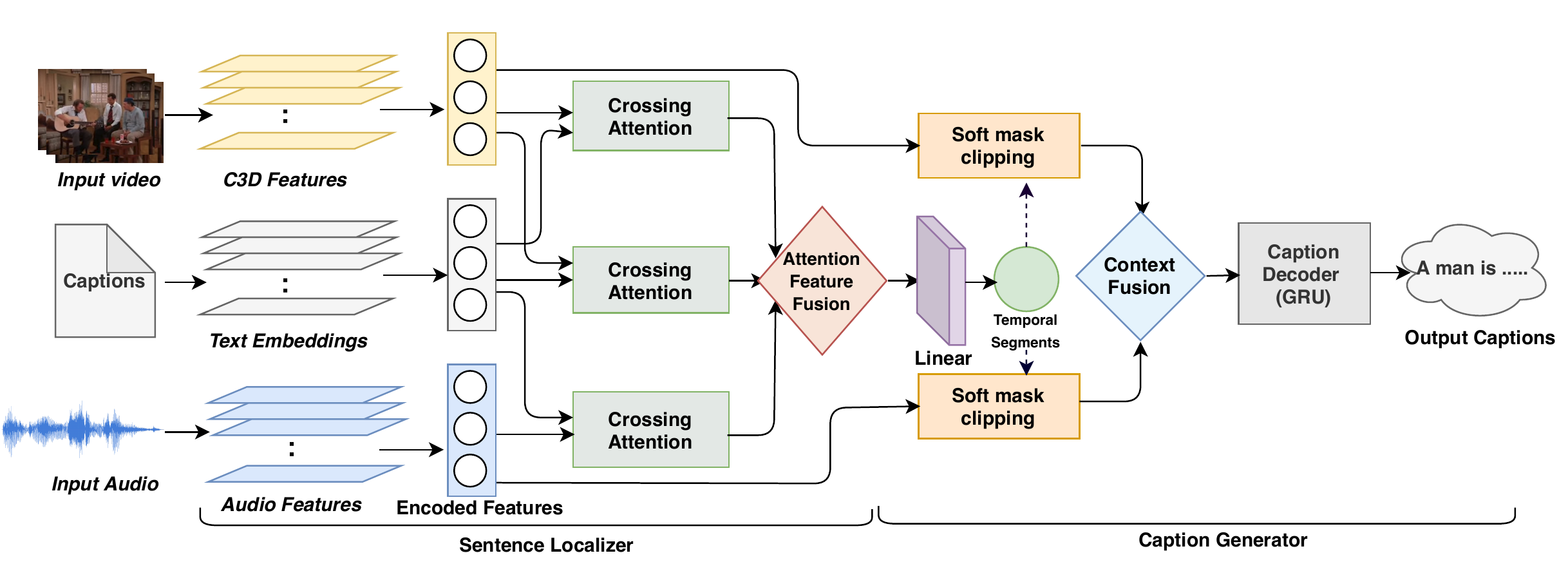}
\caption{
{\bf Our Multi-modal Architecture.} The model has two parts, a {\em sentence localizer} and a {\em caption generator}. The {\em sentence localizer} takes audio, video, and captions as inputs and generates a temporal segment for each caption. The {\em caption generator} uses the resultant temporal segments, with audio and video features, to produce a caption for each segment.}
\label{fig:block_diagram}
\vspace{-0.22in}
\end{figure*}

In this work, we consider two important modalities, audio and video, to generate dense captions in a weakly supervised setting. Weak supervision means that we do not require ground-truth temporal event segments during training. The overview of our multi-modal architecture is shown in Figure~\ref{fig:block_diagram}. The architecture consists of two modules, a {\em sentence localizer} and a {\em caption generator}. 
Given a set of initial random proposal segments in a video, caption generator produces captions for the specified segments. Sentence localizer then refines the corresponding segments with the generated captions. Caption generator is employed again to refine the captions. This process can proceed iteratively to arrive at consistent segments and captions; in practice we use one iteration following the observations in~\cite{duan2018weakly}. 



We extract features from audio, video, and captions first, and pass them as inputs to the sentence localizer during training. For each modality, an encoder is used to encode the input. We use recurrent neural networks (RNNs) with GRU~\cite{cho2014learning} units as encoders. We then apply a \textit{crossing attention} among the audio, video and caption features. Then an \textit{attention feature fusion} mechanism followed by a fully-connected layer is applied to produce temporal segments. 

The caption generator takes the encoded features of audio and video, along with the resultant temporal segments as inputs. It performs~\textit{soft mask clipping} on the audio and video features based on the temporal segments, and uses a \textit{context fusion} technique to generate the multi-modal context features. Then a caption decoder, which is also an RNN with GRU units, generates one caption for each multi-modal context feature. We discuss and compare three different context fusion strategies to find the most appropriate one for our multi-modal integration. 


In what follows, we first describe how to extract features from audio and video in Sec.~\ref{Feature_Representation}. Then we present our weakly supervised approach in Sec.~\ref{wsdec}. Lastly, we demonstrate three different context fusion strategies in section~\ref{cf}.

\subsection{Feature Representation}\label{Feature_Representation}
We consider both features from audio and video modalities for dense event captioning. It is generally challenging to select the most appropriate feature extraction process, especially for the audio modality. We describe different feature extraction methods to process both audio and video inputs.

\vspace{-0.20in}

\subsubsection{Audio Feature Processing} \label{sec:audio_feature}
ActivityNet Captions dataset~\cite{krishna2017dense} does not provide audio tracks. As such, we collected all audio data from the YouTube videos via the original URLs. Some videos are no longer available on YouTube. In total, we were able to collect around 15,840 audio tracks corresponding to ActivityNet videos. To process the audio, we consider and compare three different audio feature representations. 

\vspace{0.05in}
\noindent
\textbf{MFCC Features.} Mel-Frequency  Cepstrum (MFC) is a common representation for sound in digital signal processing. Mel-Frequency Cepstral Coefficients (MFCCs) are coefficients that collectively make up an MFC -- a representation of the short-term power spectrum of sound~\cite{jiang2009music}. We down-sample the audio from 44 kHz to 16 kHz and use 25 as the sampling rate. We choose 128 
MFCC features, 
with 2048 as the FFT window size and 512 as the number of samples between successive frames (\ie, hop length).

\vspace{0.05in}
\noindent
\textbf{CQT Features.} The Constant-Q-Transform (CQT) is a time-frequency representation where the frequency bins are geometrically spaced and the ratios of the center frequencies to bandwidths (Q-factors) of all bins are equal~\cite{brown1991calculation}. CQT is motivated from the human auditory system and the fundamental frequencies of the tones in Western music~\cite{schorkhuber2010constant}. We perform feature extraction by choosing 64 Hz and 60 as the minimum frequency and the number of frequency bins respectively. Similar to the MFCC features described above, we use 2048 as the FFT window size and 512 as the hop length. We use VGG-16~\cite{simonyan2014very} without the last classification layer to convert both MFCC and CQT features into 512-dimensional representations.

\vspace{0.05in}
\noindent
\textbf{SoundNet Features.} SoundNet~\cite{aytar2016soundnet} is a CNN that learns to represent raw audio waveforms. The acoustic representation is learned using two-million videos with their accompanying audios; leveraging the natural synchronization between them. We use a pretrained SoundNet~\cite{aytar2016soundnet} model to extract the 1000-dimension audio features from the 8-th convolutional layer (\ie, conv8) for each video's audio track.

\vspace{-0.10in} 

\subsubsection{Video Feature Processing}\label{sec:video_feature}
Given an input video $\mathbf{V}=\{{\mathbf{v}_t\}^{T_v}_{t=1}}$, where $\mathbf{v}_t$ is the video frame at time $t$ and $T_v$ is the video length, a 3D-CNN model is used to process the input video frames into a sequence of visual features $\{\mathbf{f}_t = F(\mathbf{v}_t:\mathbf{v}_{t+\delta})\}_{t=1}^{T_{f}}$. Here, $\delta$ means the time resolution for each feature $\mathbf{f}_t$ and $T_f$ is the length of the feature sequence. We use features extracted from encoder $F$ provided by the ActivityNet Captions dataset~\cite{krishna2017dense}, where $F$ is the pretrained C3D~\cite{ji20133d} network with $\delta=16$ frames. The dimension of the resultant C3D features is a tensor of size $T_f \times D$, where $D=500$ and $T_f=T_v/\delta$. 

\begin{figure*}[ht]
\begin{subfigure}{.33\textwidth}
  \centering
  \includegraphics[scale=.52]{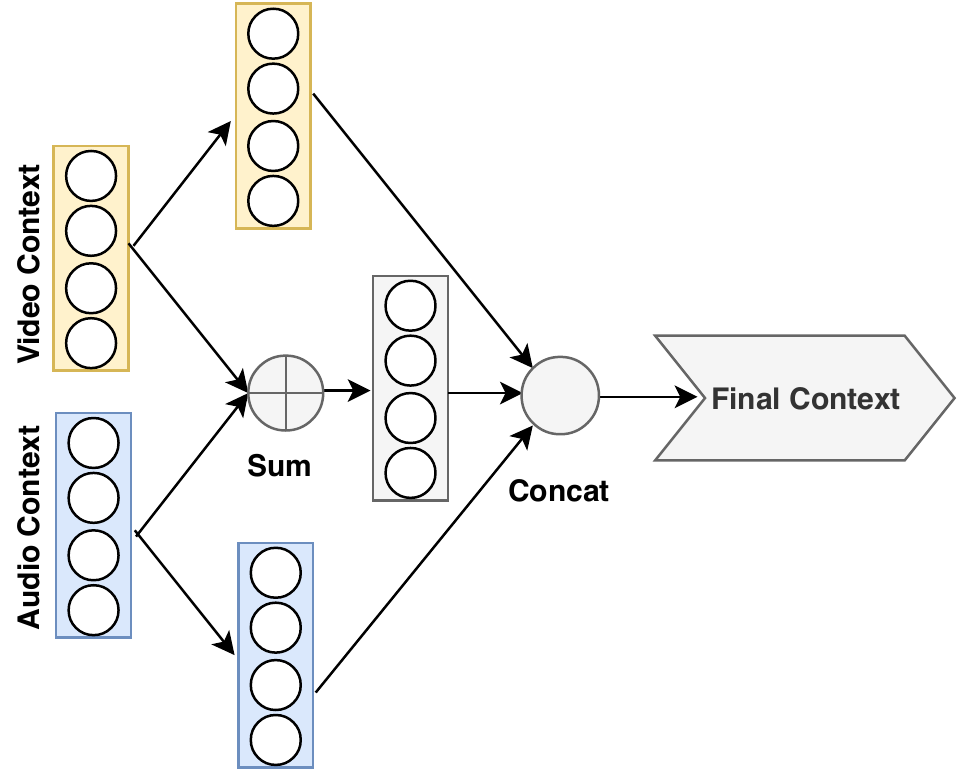}
  \caption{Multiplicative mixture fusion}
\end{subfigure}
\begin{subfigure}{.33\textwidth}
  \centering
  \includegraphics[scale=0.52]{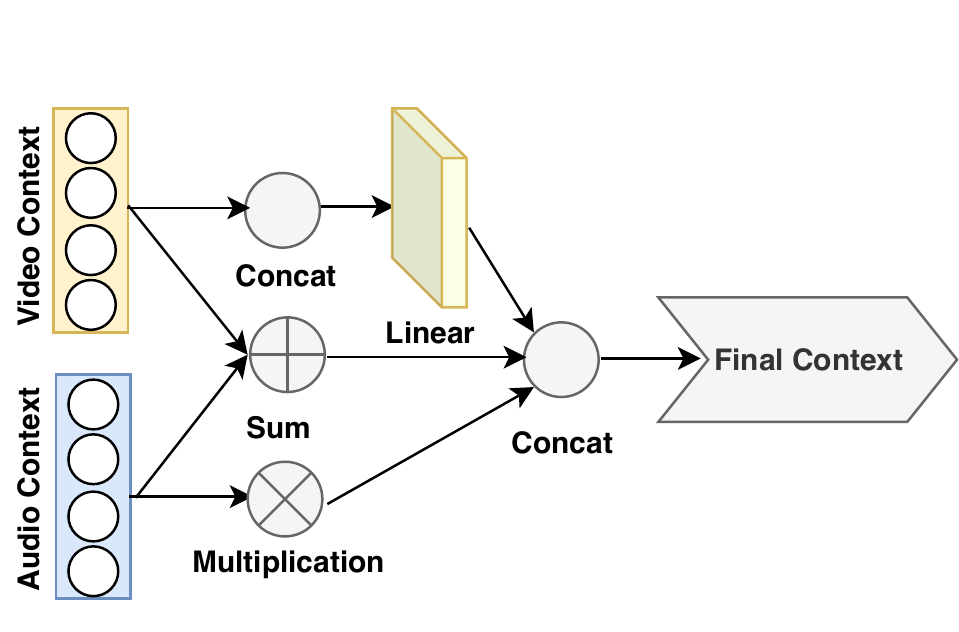}
  \caption{Multi-model context fusion}
\end{subfigure}
\begin{subfigure}{.33\textwidth}
  \centering
  \includegraphics[scale=0.52]{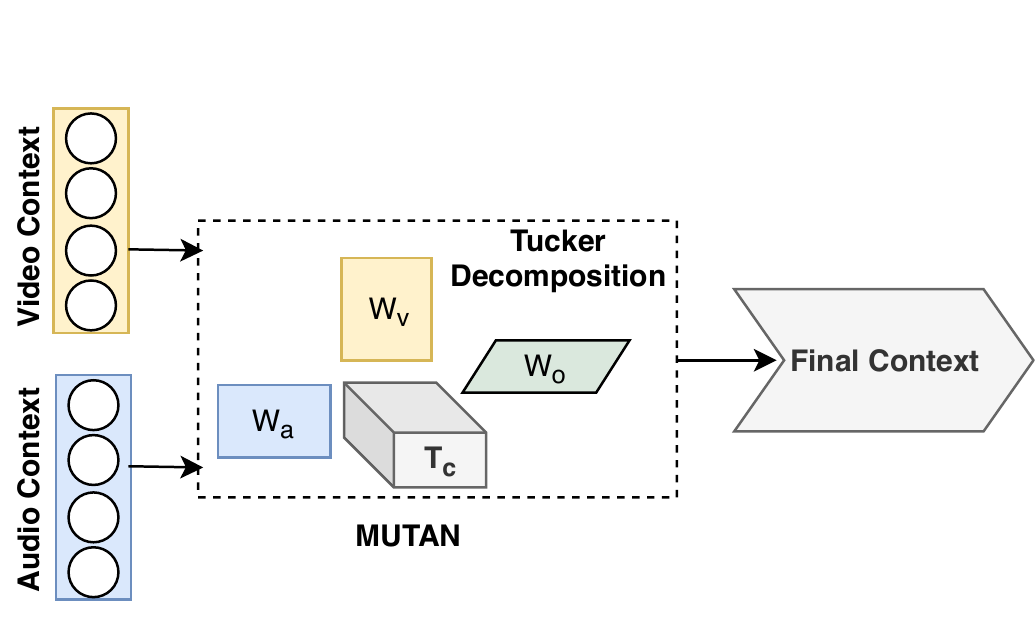}
  \caption{MUTAN fusion}
\end{subfigure}
\vspace{-0.10in}
\caption{{\bf Context Fusion Strategies.} Three fusion strategies are illustrated: (a) multiplicative mixture fusion, (b) multi-modal context fusion, and (c) MUTAN fusion.}
\label{fg:fusion}
\vspace{-0.22in}
\end{figure*}
\subsection{Weakly Supervised Model}\label{wsdec}
Weak supervision means that we do not require ground-truth temporal alignments between the video (visual and audio collectively) and captions. We make a one-to-one correspondence assumption, meaning that we assume that each caption describes one temporal segment and each temporal segment corresponds to only one caption. This assumption holds in the current benchmark dataset and most real world scenarios. We employ two network modules, a sentence localizer and a caption generator. Given a caption, the sentence localizer will produce a temporal segment in the \textit{context}, while the caption generator will generate a caption with a given temporal segment. We use \textit{context} to refer an encoded video or audio.

\vspace{0.05in}
\noindent
\textbf{Notations.}
We use GRU RNNs to encode visual and audio streams of the video. This results in a sequence of output feature vectors, one per frame, $\mathbf{O} = \{\mathbf{o}_t \in \mathbb{R}^k\}_{t=0}^{T_o}$ and the final hidden state $\mathbf{h}^{o} \in \mathbb{R}^k$, where $T_o$ is the length of the video. While in practice we get two sets of such vectors (one set for video and one set for corresponding audio ``frames"), we omit the subscript for clarity of formulation that follows. A caption is encoded similarly by the output features of the RNN, $\mathbf{C} = \{ \mathbf{c}_t \in \mathbb{R}^k\}_{t=0}^{T_c}$ with the last hidden state being $\mathbf{h}^{c} \in \mathbb{R}^k$, where $T_c$ is the length of the caption in words. We use {\em context} to refer the encoding of the full visual or audio information in videos.  
A context segment $\mathbf{S}$ is represented by $(c, l)$, where $c$ and $l$ denote segment's temporal center and length respectively within $\mathbf{O}$.


\vspace{-0.15in}
\subsubsection{Sentence Localizer}\label{sec:senlocal}
Sentence localizer attempts to localize a given caption in a video by considering the caption and the encoded complete video ({\em context}). Formally, given a (video or audio) context $\mathbf{O}$ and an encoded caption $\mathbf{C}$, sentence localizer will regress a temporal segment $\mathbf{S}$ in $\mathbf{O}$. With the context and caption features, it first applies \textit{crossing attention} among them. Then an \textit{attention feature fusion}, followed by one layer fully-connected neural network, is used to generate the temporal segment. Following ~\cite{duan2018weakly}, we use 15 predefined temporal segments and generate 15 offsets in sentence localization using fully connected layer. The final segments are the sum of temporal segments and offsets value. The purpose is to fine-tune the offset value for best localization.

\vspace{0.05in}
\noindent
\textbf{Crossing Attention.} The crossing attention consists of two sub-attentions, one caption attention $\mathbf{Att}_\mathrm{c}$, and one context attention $\mathbf{Att}_\mathrm{o}$. For a context $\mathbf{O}$ and a caption $\mathbf{C}$, we first compute the attention between $\mathbf{h}^{o}$ and $\mathbf{C}$ as: 
\begin{equation} \label{eq:att_c}
    \mathbf{Att}_\mathrm{c} = \operatorname{softmax}((\mathbf{h}^o)^T\bm{\alpha}_c\mathbf{C})\mathbf{C}^T,
\end{equation}
and then calculate the attention between $\mathbf{h}^{c}$ and $\mathbf{O}$ as:
\begin{equation} \label{eq:att_o}
    \mathbf{Att}_\mathrm{o} = \operatorname{softmax}((\mathbf{h}^c)^T\bm{\alpha}_o\mathbf{O})\mathbf{O}^T,
\end{equation}
where $\bm{\alpha}_c \in \mathbb{R}^{k \times k}$ and $\bm{\alpha}_o \in \mathbb{R}^{k \times k}$ are the learnable attention weights, and $()^T$ is the matrix transpose operation. We note that $\mathbf{Att}_\mathrm{o}$ is a vector of size $1 \times k$ comprising of attention weighted features for the visual/audio frames; similarly $\mathbf{Att}_\mathrm{c}$ is a vector of size $1 \times k$ of attended caption features. 

When training our multi-modal approaches, the caption attention $\mathbf{Att}_\mathrm{c}$ is calculated only between the visual modality and the captions, and we generate video attention $\mathbf{Att}_\mathrm{v}$ and audio attention $\mathbf{Att}_\mathrm{a}$ using Eq.~\ref{eq:att_o}. While we are training our unimodal approaches which either use audio (or video) information to generate captions, the caption attention $\mathbf{Att}_\mathrm{c}$ is calculated between the audio (or video) and captions.

\vspace{0.05in}
\noindent
\textbf{Attention Feature Fusion.} After obtaining the sub-attentions, we use the multi-model feature fusion technique~\cite{gao2017tall} to fuse them together:
\begin{align}
    \mathbf{Att}_{\mathrm{sum}} &= \mathbf{Att}_\mathrm{c} + \mathbf{Att}_\mathrm{v} + \mathbf{Att}_\mathrm{a} \\ 
    \mathbf{Att}_{\mathrm{dot}} &= \mathbf{Att}_\mathrm{c} \cdot \mathbf{Att}_\mathrm{v} \cdot \mathbf{Att}_\mathrm{a} \\ 
    \mathbf{Att}_{\mathrm{fc}} &= \operatorname{fc}(\mathbf{Att}_\mathrm{c} || \mathbf{Att}_\mathrm{v} || \mathbf{Att}_\mathrm{a}) \\ 
    \mathbf{Att}_{\mathrm{fusion}} &= \mathbf{Att}_{\mathrm{sum}}||\mathbf{Att}_{\mathrm{dot}}||\mathbf{Att}_{\mathrm{fc}} \label{feature_fusion}
\end{align}
where $+$ and $\cdot$ are the element-wise addition and multiplication, $||$ is the column-wise concatenation, and $\operatorname{fc}(\cdot)$ is a one-layer fully-connected neural network.

\subsubsection{Caption Generator}
Given a temporal segment $\mathbf{S}$ in a context $\mathbf{O}$, the caption generator will generate a caption based on $\mathbf{S}$. With the temporal segments generated by the sentence localizer (Sec.~\ref{sec:senlocal}), the caption generator first applies \textit{soft mask clipping} on the contexts, and then uses a \textit{context fusion} mechanism (Sec.~\ref{cf}) to fuse the clipped contexts together. The fused contexts are then fed to a caption decoder, which is also a GRU RNN, to generate the corresponding captions. 

\vspace{0.05in}
\noindent
\textbf{Soft Mask Clipping.} Getting a temporal segment $\mathbf{S}$ from a context, \ie, the clipping operation, is non-differentiable, which makes it difficult to handle in end-to-end training. To this end, we utilize a continuous mask function with regard to the time step $t$ to perform soft clipping. The mask $\operatorname{M}$ to obtain an $\mathbf{S}$ is defined as follows:
\begin{equation}
    \operatorname{M}(t, \mathbf{S}) = \operatorname{\sigma}(-L(t-c+\frac{l}{2})) - \operatorname{\sigma}(-L(t-c-\frac{l}{2})),
\end{equation}
where $\operatorname{\sigma}(\cdot)$ is the sigmoid function, and $L$ is a scaling factor. When $L$ is large enough, this mask function becomes a step function which performs the exact clipping. We use the normalized weighted sum of the context features (weighted by the mask) as a feature representing $\mathbf{S}$. This operation approximates traditional mean-pooling over clipped frames. 


\subsection{Context Fusion}\label{cf}
Because audio and visual representations are from two different modalities, merging them together is a crucial task in a multi-modal setting. We use three different context merging techniques (Fig.~\ref{fg:fusion}) to fuse the video $\mathbf{V}^\prime$ and audio $\mathbf{A}^\prime$ features obtained after the normalized soft mask clipping operation. We treat $\mathbf{V}^{\prime}$ and $\mathbf{A}^\prime$ as row vectors.

\vspace{0.05in}
\noindent
\textbf{Multiplicative Mixture Fusion.} The multiplicative mixture fusion can make the model automatically focus on information from a more reliable modality and reduce emphasis on the less reliable one~\cite{liu2018learn}. Given a pair of features $\mathbf{V}^{\prime}$ and $\mathbf{A}^\prime$, the multiplicative mixture fusion first adds these two contexts and then concatenates the added context with the two original ones. That is, it produces a final context as follows,
\begin{equation}
    \mathbf{C}_{\mathrm{final}} = (\mathbf{V}^{\prime} + \mathbf{A}^\prime) || \mathbf{V}^{\prime} || \mathbf{A}^\prime
\end{equation}
where $+$ and $||$ are the element-wise addition and column-wise concatenation respectively.

\vspace{0.05in}
\noindent
\textbf{Multi-modal Context Fusion.} This fusion strategy is similar to Eq.~\ref{feature_fusion}. But here, we apply the fusion technique on $\mathbf{A}^\prime$ and $\mathbf{V}^{\prime}$ (segments as opposed to full video context),
\begin{equation}
    \mathbf{C}_{\mathrm{final}} = (\mathbf{V}^{\prime} + \mathbf{A}^\prime)||(\mathbf{V}^{\prime} \cdot \mathbf{A}^\prime)||\operatorname{fc}(\mathbf{V}^{\prime} || \mathbf{A}^\prime).
\end{equation}

\vspace{0.05in}
\noindent
\textbf{MUTAN Fusion.} MUTAN fusion was first proposed in~\cite{ben2017mutan} to solve visual question answering tasks by fusing visual and linguistic features. We adopt the fusion scheme to fuse $\mathbf{V}^{\prime}$ and $\mathbf{A}^\prime$. With the idea of Tucker decomposition~\cite{tucker1966some}, we first reduce the dimension of $\mathbf{V}^{\prime}$ and $\mathbf{A}^\prime$, 
\begin{align}
    \mathbf{V}'' &= \operatorname{tanh}(\mathbf{V}^{\prime} \times \mathbf{W}_v) \\
    \mathbf{A}'' &= \operatorname{tanh}(\mathbf{A}^{\prime} \times \mathbf{W}_a)
\end{align}
where $\mathbf{W}_v$ and $\mathbf{W}_a$ are learnable parameters and $\operatorname{tanh}(\cdot)$ is the hyperbolic tangent function. Then we produce final context as folows:
\begin{align}
     \tilde{\mathbf{C}} &= ((\mathbf{T}_c \times_1 \mathbf{V}'') \times_2 \mathbf{A}'') \\
     \mathbf{C}_{\mathrm{final}} &=   \operatorname{squeeze}(\tilde{\mathbf{C}}) \times \mathbf{W}_o,
\end{align}
where $\mathbf{T}_c$ and $\mathbf{W}_o$ are learnable parameters. $\times_i, i \in \{1, 2\}$ denotes the mode-$i$ product between a tensor and a matrix, and $\times$ is the matrix multiplication operation. $\mathbf{T}_c$ models the interactions between the video and the audio modalities, which is a 3-dimension tensor; $\operatorname{squeeze}$ operator squeezes $\mathbf{\tilde{C}}$ into a row vector. 

\subsection{Training Loss}
We follow the training procedure and loss function presented in~\cite{duan2018weakly} to train our networks. We employ the idea of cycle consistency~\cite{zhu2017unpaired} to train the sentence localizer and the caption generator, and treat the temporal segment regression as a classification problem. The final training loss is formulated as
\begin{equation}\label{eq:loss}
    L =  L_{\mathrm{c}} + \lambda_{s}L_{\mathrm{s}} + \lambda_{r}L_{\mathrm{r}}
\end{equation}
where $\lambda_{s}$ and $\lambda_{r}$ are tunable hyperpramaters. $L_{\mathrm{c}}$ is the caption reconstruction loss, which is a cross-entropy loss measuring the similarity between two sentences. $L_{\mathrm{s}}$ is the segment reconstruction loss, which is an L2 loss. It measures the similarity between two temporal segments. $L_{\mathrm{r}}$ is the temporal segment regression loss, which is also a cross-entropy loss, because we regard the temporal segment regression as a classification problem.

\section{Experiments}
In this section, we first describe the dataset used in our experiments, which is an extension of the ActivityNet Captions Dataset~\cite{krishna2017dense} (Sec.~\ref{sec:dataset}). Then we present the experimental setup and implementation details (Sec.~\ref{sec:setup}). Lastly, we discuss the experimental results for both unimodal (\ie, trained using either audio or video modality) and multi-modal approaches (Sec.~\ref{sec:results}).

\subsection{Dataset}\label{sec:dataset}
ActivityNet Captions dataset~\cite{krishna2017dense} is a benchmark for large-scale dense event captioning in videos. The dataset consists of 20,000 videos where each video is annotated with a series of temporally aligned captions. On average, one video corresponds to 3.65 captions. However, besides the captions, the current dataset only provides C3D features~\cite{ji20133d} for visual frames, no original videos. To obtain the audio tracks for those videos, we needed to find the original videos on YouTube and download the audios via the provided URLs. Around 5,000 videos are unavailable on YouTube now. We are able to find 8026 videos (out of 10009 videos) for training and 3880 videos (out of 4917 videos) for validation. We use those available training/validation videos throughout our experiments.

\subsection{Experiment Setup and Implementation Details}\label{sec:setup}
We follow the experiment protocol in~\cite{duan2018weakly} to train and evaluate all the models. We consider the models proposed in~\cite{duan2018weakly} as our baselines, \ie, unimodal models that only utilize audio or visual features. Due to the difference in the number of videos for training and validation from the original dataset, we run all the experiments from scratch using the PyTorch implementation provided by~\cite{duan2018weakly}\footnote{\url{https://github.com/XgDuan/WSDEC}}. The dimensions of the hidden and output layers for all GRU RNNs (audio/video/caption encoders and caption decoders) are set to 512. We also follow~\cite{duan2018weakly} to build the word vocabulary (containing 6,000 words) and preprocess the words.

\begin{table}
\scriptsize
   \begin{center}
  \begin{tabular}{lcccccccc}
  \hline
    Features & M & C & R & B@1 & B@2 & B@3 & B@4 & S \\
    \hline \hline
    \multicolumn{8}{l}{\textbf{Pretrained model}} \\
    MFCC & 2.70 & 6.46 & 6.74 & 5.52 & 1.74 & 0.67 & 0.21 & 3.51\\
    CQT      & 2.38 & 5.60 & 5.72 & 4.37 & 1.57 & 0.46 & 0.13 & 2.90\\
    SoundNet & 2.63 & 5.76 & 6.99 & 6.28 & 1.81 & 0.38 & 0.12 & 3.44\\
    \hline
     \multicolumn{8}{l}{\textbf{Final model}} \\
    MFCC & 3.36 & 9.56 & 8.51 & 6.68 & 2.55 & 1.23 & 0.60 & 4.20\\
    CQT & 3.25 & 8.97 & 7.43 & 6.34 & 2.69 & 0.93 & 0.32 & 3.63\\
    SoundNet & 3.41 & 9.21 & 8.50 & 7.19 & 2.15 & 0.49 & 0.13 & 4.22 \\
    \hline
  \end{tabular}
  \end{center}
  \vspace{-0.15in}
  \caption{{\bf Audio Only Results.} Illustrated are dense captioning results of pretrained and final models using audio only.}
  \label{table:pretrain_cg_audio}
  \vspace{-0.22in}
\end{table}

\begin{table*}
\footnotesize
   \begin{center}
  \begin{tabular}{lccccccccc}
  \hline
    Fusion Strategies & M & C & R & B@1 & B@2 & B@3 & B@4 & S & mIoU\\
    \hline \hline
    \multicolumn{8}{l}{\textbf{Pretrained model}} \\
    Multiplicative mixture fusion & 3.59 & 8.12 & 7.51 & 7.12 & 2.74 & 1.22 & 0.56 & 4.58 & -\\
    Multi-modal context fusion & 3.55 & 7.91 & 7.54 & 7.24 & 2.78 & 1.28 & 0.62 & 4.45 & -\\
    MUTAN fusion & \textbf{3.71} & \textbf{8.20} & \textbf{7.71} & \textbf{7.45} & \textbf{2.92} & \textbf{1.31} & \textbf{0.63} & \textbf{4.78} & -\\
    \hline
     \multicolumn{8}{l}{\textbf{Final model}} \\
     Multiplicative mixture fusion & 4.89 & \textbf{13.97} & 10.39 & 9.92 & 4.17 & 1.85 & 0.88 & 5.95 & 29.87\\
     Multi-modal context fusion & \textbf{4.94} & 13.90 & 10.37 & 9.95 & 4.20 & \textbf{1.86} & 0.89 & 5.98 & 29.91\\
     MUTAN fusion & 4.93 & 13.79 & \textbf{10.39} & \textbf{10.00} & \textbf{4.20} & 1.85 & \textbf{0.90} & \textbf{6.01} & \textbf{30.02}\\
    \hline
  \end{tabular}
  \end{center}
  \vspace{-0.20in}
  \caption{{\bf Fusion Strategies.} Testing results for different context fusion strategies for integrating audio and video modalities are illustrated for both pretrained and final models. We use MFCC audio features and C3D video features for all experiments.}
  \label{table:fusion_comparison}
\end{table*}

\begin{table*}
\footnotesize
   \begin{center}
  \begin{tabular}{lcccccccccc}
  \hline
    Model & M & C & R & B@1 & B@2 & B@3 & B@4 & S & mIoU \\ 
    \hline \hline
    \multicolumn{8}{l}{\textbf{Pretrained model}} \\
    Unimodal (C3D video feature)~\cite{duan2018weakly} & 3.66 & 8.20 & 7.42 & 7.06 & 2.76 & 1.29 & 0.62 & 4.41 & - \\ 
    Unimodal (SoundNet audio feature) & 2.63 & 5.76 & 6.99 & 6.28 & 1.81 & 0.38 & 0.12 & 3.44 & - \\ 
    Unimodal (MFCC audio feature) & 2.70 & 6.46 & 6.74 & 5.52 & 1.74 & 0.67 & 0.21 & 3.51 & - \\ 
    \hline
    Multi-modal (SoundNet audio + C3D video feature) & 3.72 & 8.02 & 7.50 & 7.12 & 2.74 & 1.23 & 0.58 & 4.46 & - \\  
    Multi-modal (MFCC audio + C3D video feature) & 3.71 & 8.20 & 7.71 & 7.45 & 2.92 & 1.31 & 0.63 & 4.78 & - \\
    \hline
    \multicolumn{8}{l}{\textbf{Final model}} \\
    Unimodal (C3D video feature)~\cite{duan2018weakly} & 4.89 & 13.81 & 9.92 & 9.45 & 3.97 & 1.75 & 0.83 & 5.83 & 29.78 \\ 
    Unimodal (SoundNet audio feature) & 3.41 & 9.21 & 8.50 & 7.19 & 2.15 & 0.49 & 0.13 & 4.22 & 25.57 \\ 
    Unimodal (MFCC audio feature) & 3.36 & 9.56 & 8.51 & 6.68 & 2.55 & 1.23 & 0.60 & 4.20 & 27.16 \\  
    \hline
    Multi-modal (SoundNet audio + C3D video feature) & {\bf 5.03} & {\bf 14.27} & 10.35 & 9.75 & 4.19 & {\bf 1.92} & {\bf 0.94} & {\bf 6.04} & 29.96 \\ 
    Multi-modal (MFCC audio + C3D video feature) & 4.93 & 13.79 & {\bf 10.39} & {\bf 10.00} & {\bf 4.20} & 1.85 & 0.90 & 6.01 & {\bf 30.02} \\ 
    \hline 
  \end{tabular}
  \end{center}
  \vspace{-0.20in}
  \caption{{\bf Multi-modal Results.} Comparison among unimodal and our multi-modal models using MUTAN fusion.}
  \label{table:multimodal_results}
  \vspace{-0.10in}
\end{table*}

\begin{table*}[h]
\vspace{-0.05in}
\footnotesize
   \begin{center}
  \begin{tabular}{lcccccccc}
  \hline
    Model & M & C & R & B@1 & B@2 & B@3 & B@4 & S\\
    \hline \hline
    Unimodal (C3D)~\cite{duan2018weakly} & 7.09 & 24.46 & 14.79 & 14.32 & 6.23 & 2.89 & 1.35 & 8.22\\
    Multi-modal (SoundNet audio feature + C3D video feature) & 7.02 & 24.22 & 14.66 & 14.18 & 6.13 & 2.88 & 1.41 & 7.89\\
    Multi-modal (MFCC audio feature + C3D video feature) & \textbf{7.23} & \textbf{25.36} & \textbf{15.37} & \textbf{15.23} & \textbf{6.58} & \textbf{3.04} & \textbf{1.46} & \textbf{8.51}\\
    \hline
  \end{tabular}
  \end{center}
  \vspace{-0.20in}
  \caption{Results with ground-truth temporal segments.}
  \label{table:gt_seg_results_rebuttal}
  \vspace{-0.05in}
\end{table*}

\begin{table*}[h]
\footnotesize
   \begin{center}
  \begin{tabular}{lcccccccc}
  \hline
    Model & M & C & R & B@1 & B@2 & B@3 & B@4 & S \\
    \hline \hline
    Unimodal (C3D)~\cite{duan2018weakly} & 4.58 & 10.45 & 9.27 & 8.7 & 3.39 & 1.50 & 0.69 & -\\
    Multi-modal (SoundNet + C3D) & 4.70 & 10.32 & 9.40 & 8.95 & 3.40 & 1.53 & 0.73 & 5.51\\
    Multi-modal (MFCC + C3D) & \textbf{4.78} & \textbf{10.53} & \textbf{9.60} & \textbf{9.23} & \textbf{3.62} & \textbf{1.69} & \textbf{0.82} & \textbf{5.56}\\
    \hline
  \end{tabular}
  \end{center}
  \vspace{-0.20in}
  \caption{Pretrained model results on the full dataset.}
  \label{table:multimodal_results_rebuttal}
  \vspace{-0.20in}
\end{table*}

\begin{figure*}[h!]
  \centering
  \includegraphics[scale=0.78]{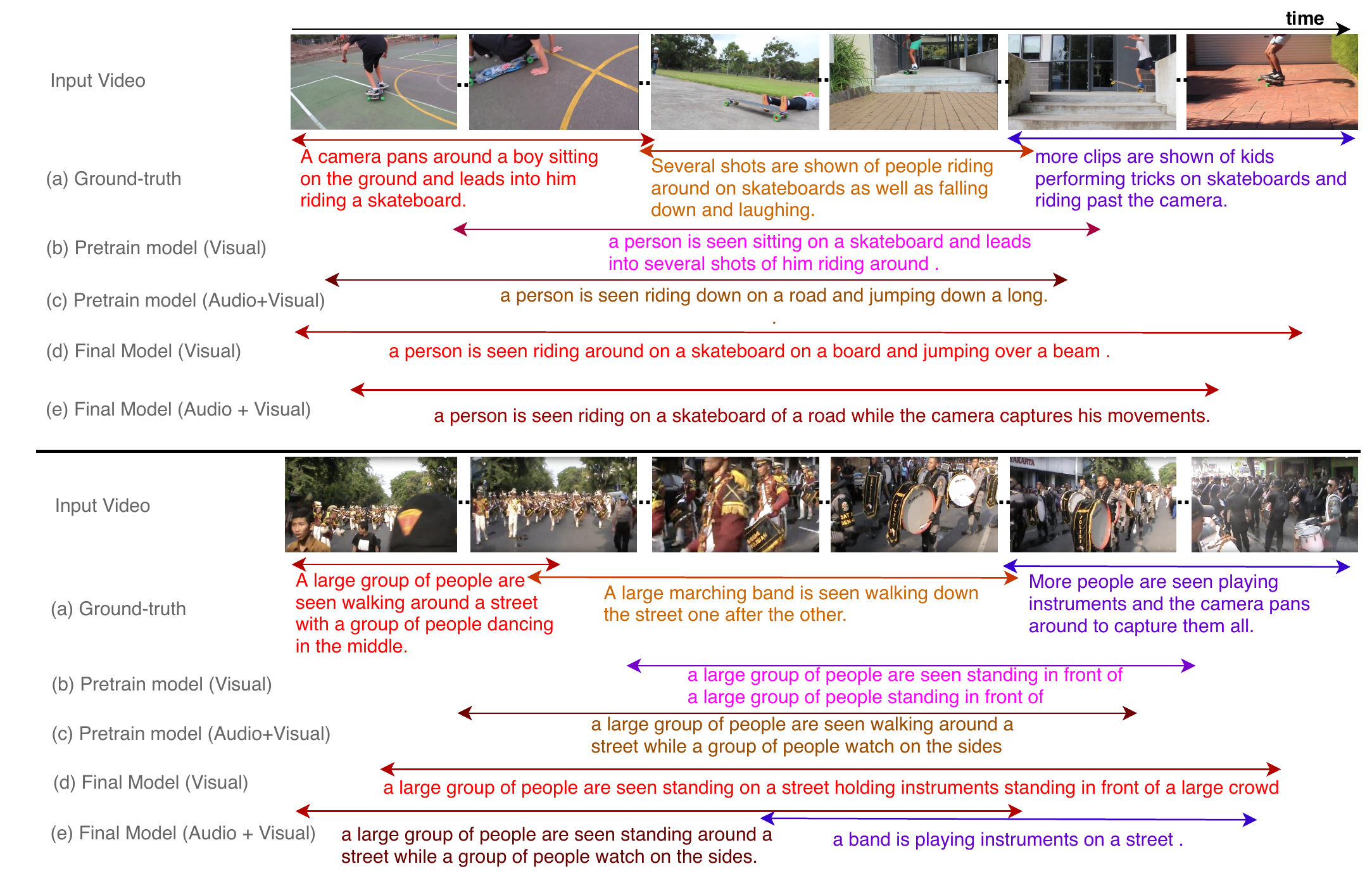}
\caption{{\bf Qualitative Results.} Both pretrained and final model results are illustrated of two videos. Captions are from (a) ground-truth; (b) pretrained model trained only using visual features; (c) multi-modal pretrained model; (d) final model trained with video features only; (e) our multi-modal final model for dense event captioning in videos.}
\label{fig:qualatative_result}
\vspace{-0.20in}
\end{figure*}

\vspace{0.05in}
\noindent
\textbf{Training.} Weak supervision means that we do not have ground-truth temporal segments. We first train the caption generator only (\textit{pretrained model}), and then train the sentence localizer and caption generator together (\textit{final model}). To train the pretrained model, we input the entire context sequence (\textit{Fake Proposal}, $\mathbf{S}=(0.5, 1)$). We use the weights of the pretrained model to initialize the relevant weights in the final model. 
For both pretrained model and final model, we train them in both unimodal and multi-modal settings.
To train unimodal models, we use initial learning rates 0.0001 and 0.01 for audio and video respectively with a cross-entropy loss. While training our multi-modal models, we set the initial learning rates to 0.0001 for the network parts that have been initialized with the pretrained weights, and 0.01 for other network components. $\lambda_s$ and $\lambda_r$ in Eq.~\ref{eq:loss} are both set to 0.1. We train the networks using stochastic gradient descent with a momentum factor of 0.8. 

\vspace{0.05in}
\noindent
\textbf{Testing.} 
To test the pretrained models, we select one random ground truth description as well as random temporal segment instead of entire video unlike training.
For the final models, following~\cite{duan2018weakly}, we start from 15 randomly guessed temporal segments, and apply one round of fixed-point iteration and the IoU filtering mechanism to obtain a set of filtered segments. Caption generators are applied to the filtered segments together with context features to produce the dense event captions. 

\vspace{0.05in}
\noindent
\textbf{Evaluation metrics.} We measure the performance of captioning results using traditional evaluation metrics: METEOR (M)~\cite{banerjee2005meteor}, CIDEr (C)~\cite{vedantam2015cider}, Rouge-L (R)~\cite{lin2004automatic}, Spice (S)~\cite{anderson2016spice} and Bleu@N (B@N)~\cite{papineni2002bleu}. For score computations, we use official scripts provided by~\cite{krishna2017dense}\footnote{\url{https://github.com/ranjaykrishna/densevid_eval}}. Where appropriate, we use mean Intesection over Union (mIoU) to measure segment localization performance.

\subsection{Experiment Results}\label{sec:results}
Since audio features can be represented in a variety of ways~\cite{aytar2016soundnet, schorkhuber2010constant, vergin1999generalized}, finding the best representation is challenging. We conduct experiments on both pretrained models and final models using different audio representations, \ie, MFCC~\cite{jiang2009music}, CQT~\cite{brown1991calculation}, and SoundNet~\cite{aytar2016soundnet}, which are described in Sec.~\ref{sec:audio_feature}. Table~\ref{table:pretrain_cg_audio} shows the experiment results of pretrained models and final models using only audio features. We can see that both MFCC and soundNet can generate comparable results. 

As discussed in Sec.~\ref{cf}, in the multi-modal setting, choosing a good fusion strategy to combine both audio and video features is another crucial point. Table~\ref{table:fusion_comparison} shows comparison of different context fusion techniques using MFCC audio representations and C3D visual features (Sec.~\ref{sec:video_feature}) for both pretrained models and final models. Among all fusion techniques, we find that MUTAN fusion is the most appropriate one for our weakly supervised multi-modal dense event captioning task. Therefore, we decide to use MUTAN fusion technique for our multi-modal models when comparing to unimodal models. Tab.~\ref{table:multimodal_results} shows the testing results for comparison among unimodal and multi-modal approaches. We can see that our multi-modal approach (both MFCC and SoundNet audio with C3D video features) outperforms state-of-the-art unimodal method~\cite{duan2018weakly} in most evaluation metrics. Specifically on the Bleu@3 and Bleu@4 scores, it leads to 9\% and 13\% improvement respectively. Comparing among unimodel approaches, we are surprised to find that only using audio features achieves competitive performance.
We trained our caption generator with GT segments to remove the effect of localization. The results are shown in Table~\ref{table:gt_seg_results_rebuttal}. We also conduct experiment on pretrain caption generator using the full dataset where for some videos, audio data is not available (treated as missing data). We consider zero feature vectors for missing audios. The results are shown in Table~\ref{table:multimodal_results_rebuttal}. In addition, we randomly selected 15 validation videos and invited 20 people to conduct human evaluation for comparing our multi-modal model to the visual-only one. The forced choice preference rate for our multi-modal model is 60.67\%.

Figure~\ref{fig:qualatative_result} demonstrates some qualitative results for both pretrained models and final models. It displays the ground-truth captions along with the ones generated by unimodal models and our multi-modal models. The arrow segments indicate the ground-truth or detected temporal event segments. We utilize C3D visual features along with audio features. We can see that our multi-modal approaches outperform unimodal ones, both on caption quality and temporal segment accuracy.

Similar to~\cite{duan2018weakly}, we are suffering from two limitations. One is that sometimes our multi-modal model can not detect the beginning of an event correctly. The other is that most of the time our final model only generates around 2 event captions, which means that the multi-modal approach is still not good enough to detect all the events in the weakly supervised setting. Overcoming of these two limitations is the focus of our future work.

\vspace{-0.20in}

\section{Conclusion}
\vspace{-0.08in}
Audio is a less explored modality in the computer vision community. In this paper, we propose a muti-modal approach for dense event captioning in a weakly supervised setting. We incorporate both audio features with visual ones to generate dense event captions for given videos. We discuss and compare different feature representation methods and context fusion strategies. Extensive experiments illustrate that audio features can play a vital role, and combining both audio and visual modalities can achieve performance better than the state-of-the-art unimodal visual model. 

\vspace{0.05in}
\noindent
\textbf{Acknowledgments:} This work was funded in part by the Vector Institute for AI, Canada CIFAR AI Chair, NSERC Canada Research Chair (CRC) and an NSERC Discovery and Discovery Accelerator Supplement Grants. 

{\small
\bibliographystyle{ieee_fullname}
\bibliography{egbib}
}

\end{document}